\documentclass[double]{article}
\usepackage{times}
\usepackage{algorithm}
\usepackage{algorithmic}
\usepackage{wrapfig}
\usepackage{verbatim}
\usepackage{amsmath}
\usepackage{graphicx}
\usepackage{subcaption}
\usepackage[active]{srcltx}
\usepackage{color}
\usepackage{lineno}
\usepackage{dirtytalk}
\usepackage{amsthm}
\usepackage{authblk}
\usepackage{listings}
\usepackage{tikz}
\usetikzlibrary{shapes.geometric, arrows}
\usepackage{threeparttable}
\usepackage{xurl}

\usepackage{epsfig,graphicx,amsfonts}
\usepackage{amsmath,amsthm,amssymb}
\usepackage{xcolor}

\usepackage{adjustbox}
\usepackage{multirow}
\usepackage{setspace}
\usepackage{hyperref}
\usepackage{comment}

\long\def\comment #1\commentend{}

\begin{document}

\title{\Large Symbolic Regression as Feature Engineering Method for Machine and Deep Learning Regression Tasks}

\author{Assaf Shmuel$^{1,*}$, Oren Glickman$^{1}$, Teddy Lazebnik$^{2,3}$\\
\(^1\) Department of Computer Science, Bar Ilan University, Ramat Gan, Israel\\
\(^2\) Department of Mathematics, Ariel University, Ariel, Israel\\
\(^3\) Department of Cancer Biology, Cancer Institute, University College London, London, UK\\
\(^*\) Corresponding author: assafshmuel91@gmail.com

}

\date{ }

\maketitle 

\begin{abstract}
In the realm of machine and deep learning regression tasks, the role of effective feature engineering (FE) is pivotal in enhancing model performance. Traditional approaches of FE often rely on domain expertise to manually design features for machine learning models. In the context of deep learning models, the FE is embedded in the neural network's architecture, making it hard for interpretation. In this study, we propose to integrate symbolic regression (SR) as an FE process before a machine learning model to improve its performance. We show, through extensive experimentation on synthetic and real-world physics-related datasets, that the incorporation of SR-derived features significantly enhances the predictive capabilities of both machine and deep learning regression models with 34-86\% root mean square error (RMSE) improvement in synthetic datasets and 4-11.5\% improvement in real-world datasets. In addition, as a realistic use-case, we show the proposed method improves the machine learning performance in predicting superconducting critical temperatures based on Eliashberg theory by more than 20\% in terms of RMSE. These results outline the potential of SR as an FE component in data-driven models.  \\ \\
\noindent
\textbf{Keywords:} symbolic regression; neural network; data-driven physics; feature engineering; data science
\end{abstract}

\maketitle \thispagestyle{empty}

\pagestyle{myheadings} \markboth{Draft:  \today}{Draft:  \today}
\setcounter{page}{1}

\section{Introduction}
\label{sec:introduction}

Machine and deep learning, achieving optimal performance and interpretability in regression tasks stands as a fundamental computational challenge, critical for applications spanning various fields of science and engineering \cite{kutz2017deep,reichstein2019deep,alzubaidi2021review,ode_sr_2,virgolin2020machine,teddy_3,teddy_4}. The efficacy of the machine learning  (ML) pipeline hinges on numerous components that govern its performance \cite{zhong2005comparison,discussion_1,he2021automl}, explainability \cite{exp_1,exp_2}, and development efficiency \cite{li2018ease,dalessandro2013bring}. Arguably, feature selection and feature engineering (FE) are the most important steps \cite{chandrashekar2014survey} as they are commonly one of the first steps in the ML pipeline and therefore determine the performance of everything computed afterward \cite{miao2016survey}. Moreover, the FE process is instrumental in transforming raw data into meaningful representations that empower models to capture underlying patterns effectively \cite{f_i_1,f_i_2}. 

Traditionally, crafting relevant features necessitates domain expertise, a labor-intensive approach that may not fully harness the intricate relationships within complex datasets \cite{fs_c_1,fs_c_2,fs_c_3}. In contrast, contemporary deep learning (DL) models treat feature extraction as an inherent \say{black box} process, relinquishing some degree of interpretability and control in favor of automated extraction \cite{alzubaidi2021review,pick_tau_3}. Between these extremes, alternative FE methods aim to furnish meaningful features that improve a down-the-line objective \cite{zhu2010feature,review_wfs}. While intuitive and promising, the last group is usually the most time-consuming and computationally intensive, making it challenging or even unrealistic to be efficiently utilized for many cases \cite{review_wfs}. 

To address this challenge, this paper proposes a paradigm shift by advocating the integration of Symbolic Regression (SR) as an FE method. SR empowers models to autonomously evolve mathematical expressions, capturing intricate data relationships and uncovering latent features that elude manual construction \cite{AI_Feynman,intro_8,intro_9}. By incorporating SR-derived features before applying ML or DL models.

To evaluate this method's performance, we employ an "off-the-shelf" SR model (GPlearn \cite{gp_model}), automatic ML (TPOT \cite{tpot}), and automatic DL models (AutoKeras \cite{autokeras}) on both synthetic and real-world datasets. We investigate how the introduction of SR-derived features impacts the performance of data-driven models while reducing the reliance on developer expertise. Our results demonstrate that integrating SR-derived features significantly enhances (\(p<0.01\)) the predictive capabilities of both ML and DL regression models across a wide range of synthetic (\(n = 1250\)) and real-world (\(n = 1000\)) datasets, yielding performance gains of up to \(86.0\%\) and \(34.4\%\) for synthetic datasets and \(4.0\%\) and \(11.5\%\) for real-world datasets, respectively.

The remainder of this paper is organized as follows: Section \ref{sec:related_work} provides an overview of related work in the field of FE, discussing both traditional methods and recent advances in automated feature extraction. Section \ref{sec:methodology} delves into our proposed methodology for integrating SR into the FE process. In Section \ref{sec:experiments}, we detail our experimental setup and present the datasets used. Section \ref{sec:evolution} presents and thoroughly analyzes our results. Finally, Section \ref{sec:discussion} explores the implications and insights drawn from our findings, discussing the advantages, limitations, and potential applications of our proposed approach.

\section{Related Work}
\label{sec:related_work}
FE has long been a pivotal aspect in bolstering the performance and interpretability of ML and DL regression tasks \cite{fs_c_1,fs_c_2}. Traditional approaches have relied on the manual crafting of features by domain experts, yielding potent but labor-intensive methods \cite{AI_Feynman,intro_8,intro_9}. Conversely, recent advancements in DL have introduced automated feature extraction within the model architecture, often at the expense of interpretability \cite{alzubaidi2021review,pick_tau_3}. In this section, we provide an overview of existing research, commencing with a spectrum of FE methodologies and their implications. Subsequently, we provide an overview of the main groups of SR methods and recent developments in the field, highlighting its potential as an automated FE method.

\subsection{Feature Engineering}
\label{subsec:feature_engineering}
Informally, FE involves the creation, selection, transformation, and manipulation of features in a dataset to enhance the performance of a down-the-line data-driven model. The importance of FE in ML and regression tasks is widely acknowledged \cite{fs_important_1,fs_important_2}. As such, a growing body of work explores automated FE methods \cite{afs_1,afs_2}. Given the complexity of this task, researchers have employed diverse strategies to tackle this challenge. These include genetic programming \cite{sr_example_physics}, evolutionary algorithms \cite{e_fs}, neural architecture search for automatic feature construction \cite{he2021automl,substract}, and wrapper feature selection \cite{review_wfs}, among others. While these approaches alleviate the manual burden, they often fall short in terms of interpretability. Notably, FE often accompanies feature selection, and for readers seeking further details on feature selection, we provide a brief overview of feature selection in the Appendix.

Traditional FE methods typically necessitate domain-specific expertise to devise relevant features tailored to specific problem domains. These methods have been extensively employed in various fields, including natural language processing, computer vision, and sensor data analysis \cite{fe_example,fe_example_2,fe_case}. For instance, in the context of stroke risk prediction (as exemplified in \cite{fe_bmi}), it is common practice to incorporate a feature encompassing Body Mass Index ($\text{BMI}=\text{weight}/\text{height}^2$), effectively amalgamating weight and height parameters. 

Feature transformation is a common FE practice involving data modification to enhance modeling suitability. This may include scaling, normalization, or applying mathematical functions like logarithms to address skewed distributions. As datasets grow larger and more complex, practitioners resort to computationally intensive FE techniques. Feature extraction methods create new features by transforming or combining existing ones. Common techniques include Principal Component Analysis \cite{pca_1,pca_2}, Independent Component Analysis \cite{ica}, and autoencoders \cite{autoencoder}. These methods can automatically generate new features or reduce data dimensionality while preserving essential information, proving valuable for a wide range of ML and DL models \cite{fe_good_1,fe_good_2,fe_good_3}. For example, \cite{fe_good_example} applied Principal Component Analysis to data derived from a socioeconomic questionnaire regarding barriers to healthcare.

\subsection{Symbolic Regression}
\label{subsec:symbolic_regression}

SR is approached via various strategies, broadly classified into four primary categories based on computational techniques: brute-force search, sparse regression, DL, and genetic algorithms \cite{wang2019symbolic,la2021contemporary}. Each category has its unique merits and limitations, with no single approach prevailing over others, as demonstrated in a recent comparative survey \cite{sr_benchmark}.

Brute-force SR models theoretically possess the capability to solve any SR task by exhaustively testing all possible equations to identify the best-performing one \cite{heule2017science}. Nevertheless, the practical application of brute-force methods frequently proves unfeasible due to their substantial computational requirements, which persist as a challenge even when dealing with small datasets. Moreover, these models tend to overfit when confronted with large and noisy data \cite{riolo2013genetic}, a common scenario in many real-world datasets \cite{scimed,miller1995genetic}. DL SR models excel in handling noisy data due to neural networks' intrinsic resistance to outliers \cite{sr_example_others}. Nonetheless, empirical studies indicate limited generalization capabilities, constraining their utility \cite{sr_benchmark}. Notably, \cite{petersen2019deep} introduced a Deep SR (DSR) model catering to general SR tasks, employing reinforcement learning to train a generative Recurrent Neural Network (RNN) model for symbolic expressions. Moreover, DSR integrates a variant of the Monte Carlo policy gradient approach to customize the generative model for exact formulas. Sparse regression methods significantly narrow the exploration scope by identifying concise models through sparsity-driven optimization \cite{sr_sm_intro}. A notable algorithm tailored for scientific contexts is \textit{SINDy} \cite{brunton2016discovering}. Leveraging a Lasso linear model, \textit{SINDy} identifies sparser representations of nonlinear dynamical systems underlying time-series data. The algorithm iterates between a partial least-squares fit and a thresholding step to promote sparsity. Over time, \textit{SINDy} has undergone improvements: \cite{kaiser2018sparse} enhanced its performance with noisy data, \cite{mangan2017model} introduced optimal model selection over varying threshold values, and \cite{kaptanoglu2021pysindy} developed \textit{PySINDy}, an open-source Python package for applying \textit{SINDy}. That said, SINDy is best performed when the features are associated with physical data and it underperforms for other cases. 

Finally, genetic algorithms (GA) SR models efficiently incorporate prior knowledge to constrain the function search space \cite{kronberger2022shape}. For instance, SR can adhere to predefined solution shapes \cite{salustowicz1997probabilistic, sastry2003probabilistic, yanai2003estimation, hemberg2012investigation}, or employ probabilistic models to sample grammar rules governing solution generation \cite{shan2004grammar, bosman2004learning, wong2014grammar, sotto2017probabilistic}. An effective yet simple GA-based SR implementation is the \textit{gplearn} Python Library \cite{stephens2016genetic}. It commences by generating a population of basic random formulas, structured as tree-like relationships between independent variables (features) and dependent variables (targets). Subsequently, through stochastic optimization, it iterates on subtree replacement, recombination, fitness evaluation by executing trees, and stochastic survival of the fittest. This technique demonstrates proficiency in solving linear real-world problems \cite{la2021contemporary} and can serve as a foundational framework for more intricate models, as we present in this work.

\subsection{Automated Machine Learning}
\label{subsec:auto_ml}
AutoML (Automated Machine Learning) \cite{automl} has emerged as a transformative technology in the field of artificial intelligence and machine learning. It aims to automate and simplify the intricate process of building, training, and deploying machine learning models. Two notable frameworks within the AutoML domain are TPOT and AutoKeras, each offering unique approaches to streamline and optimize the machine learning pipeline.

TPOT (Tree-based Pipeline Optimization Tool) \cite{tpot}: TPOT is a widely recognized AutoML framework that leverages genetic programming to automatically search and construct effective machine learning pipelines. Genetic programming involves evolving a population of machine learning pipelines over generations, with each generation improving upon the previous one. TPOT explores a wide range of preprocessing techniques, feature engineering methods, and machine learning algorithms, evolving optimal combinations tailored to a specific problem. It assesses the performance of these pipelines using cross-validation and selects the best-performing model. TPOT is particularly valuable when dealing with structured data and tabular datasets.

AutoKeras \cite{autokeras}: AutoKeras is another prominent AutoML framework that specializes in the automation of deep learning model construction. It simplifies the process of neural network architecture search and hyperparameter tuning. AutoKeras employs a technique known as neural architecture search (NAS) to automatically discover the most suitable neural network architectures for a given task. This involves exploring a wide range of neural network configurations, including various layers and hyperparameters, to identify the architecture that yields the best performance. AutoKeras is especially beneficial when working with unstructured data types such as images, text, and sequences, where deep learning approaches excel.

Both TPOT and AutoKeras exemplify the power of AutoML in terms of automating complex decision-making processes in machine learning. By minimizing the need for manual intervention, they make machine learning accessible to a broader audience, accelerate model development, and enable non-experts to harness the potential of artificial intelligence for their specific applications. In a recent benchmark study \cite{Benchmarking_AutoML} AutoKeras and TPOT showed the best results for wave data classification.

\section{Symbolic Regression as Feature Engineering}
\label{sec:methodology}

In this study, our objective is to investigate the role of SR as a preliminary layer in ML and DL models, essentially functioning as a FE step, generating more intricate features for data-driven models to leverage. To achieve this, we incorporate an SR model into our workflow, introducing its output as an additional feature for subsequent ML and DL models. Formally, this technique involves the simultaneous training of two models: the SR model and the subsequent ML/DL model. The SR model is trained on the input features, denoted as \(X\), and produces a new SR-derived feature, which is appended to \(X\) to create \(X^*\). Subsequently, the ML/DL model is trained on \(X^*\). Both models address the same regression task with respect to a specified target feature, denoted as \(y\).

To be exact, let us assume a given dataset \(D \in \mathbb{R}^{n \times m}\) with \(n\) rows and \(m\) features such that the features are divided into source \((X)\) and target \((y)\) features. Consistent with common data-driven model development practices, we initially divide the rows into training and validation subsets. Focusing on the training subset to avoid data leakage \cite{Hewamalage_et_al}, we employ the well-known GPlearn SR model \cite{gp_model} to create multiple SR models for the training data, each one contributing a potential feature to \(X^*\). Each instance of the SR model is characterized by a distinct parsimony coefficient. These SR models generate multiple solutions during the search for the optimal equation. Consequently, we select the equation with the lowest Root Mean Square Error (RMSE) score across the entire search process, which may not necessarily be the latest one. Subsequently, we train either an ML model using an automatic ML model - Tree-based Pipeline Optimization Tool (TPOT) \cite{tpot} or a DL model using an automatic DL model - Autokeras (AK) \cite{autokeras}.

Once the models are trained, we compare the performance of the TPOT model on the original data \(X\) (referred to as TPOT) and the SR-enhanced data \(X^*\) (referred to as SRTPOT). Similarly, we assess the performance of the AK model on the original data \(X\) (referred to as AK) and the SR-enriched data \(X^*\) (referred to as SRAK). In both cases, we calculate the RMSE score of the TPOT or AK models divided by the score of the SRTPOT or SRAK models, respectively. We then subtract 1 from the result and multiply by 100 to express the change in percentages. Thus, in this proposed metric, values greater than zero indicate a favorable performance for the SRTPOT or SRAK models in comparison to the TPOT or AK models, respectively.

\begin{figure}
    \centering
    \includegraphics[width=0.99\textwidth]{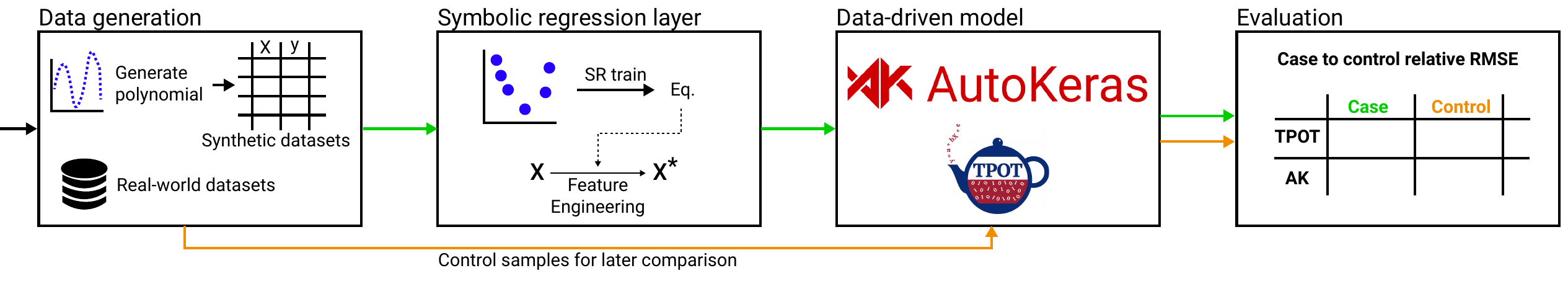}
    \caption{A schematic view of this method and experiment. We begin by either generating synthetic datasets or using real-world datasets. We then train the SR model and create the SR-based feature. Next, we train AutoML models on both the SR-enhanced data as the case sample, and the original data as control. We compare the RMSE scores in the testing data of both models.}
    \label{fig:scheme}
\end{figure}

In order to study the influence of the SR FE layer, for each dataset, we initially trained the ML/DL model on the original features (\(X\)), treating it as the control sample. Subsequently, we train the model on the SR-enhanced features (\(X^*\)), considering it as the case sample. We then compute the relative difference in RMSE between the control and case samples, ensuring the metric is unit-free and comparable across various datasets. Fig. \ref{fig:scheme} provides a schematic representation of this methodology and experiment.

Table \ref{tab:example} demonstrates the results of one  synthetic dataset. We begin by randomly generating the polynomial $x_1^3*x_2^3+x_3^3+x_1^3$. As we add 5\% Gaussian noise, the true polynomial prediction RMSE is nonzero and equals 0.29. We fit the SR model, which provides the polynomial $x_1^3*x_2^3+x_3^3+2*x_1$. While the first two terms are correct, the third term is not accurate, resulting in an RMSE score of 1.31. Even though both the TPOT and AK models outperform the SR model with RMSE scores of 0.99 and 0.70, adding the SR-derived feature substantially improves these models. When training the SRTPOT and SRAK models with the SR-derived feature, the TPOT RMSE score improves from 0.99 to 0.79 and the AK RMSE score improves from 0.70 to 0.36.

\begin{table}[ht]
\centering
\caption{Example of one synthetic dataset}
\begin{threeparttable}
\begin{tabular}{lc}
\hline
True polynomial & $x_1^3*x_2^3+x_3^3+x_1^3$ \\
\hline
Fitted SR & $x_1^3*x_2^3+x_3^3+2*x_1$ \\
\hline
Baseline – true polynomial RMSE (noise only) & 0.29 \\
\hline
LR RMSE & 9.38 \\
\hline
SR RMSE & 1.31 \\
\hline
TPOT RMSE & 0.99 \\
\hline
SRTPOT RMSE & 0.79 \\
\hline
AK RMSE & 0.70 \\
\hline
SRAK RMSE& 0.36 \\
\hline

\end{tabular}
\end{threeparttable}
\label{tab:example}
\end{table}

\section{Experimental Setup}
\label{sec:experiments}

To provide a comprehensive analysis of the proposed method, we used both synthetic and real-world datasets, focusing exclusively on regression tasks with numerical features.

\subsection{Generating Synthetic Datasets}
We initiated the experiment with synthetic datasets. To generate these datasets, we followed a procedure of randomly creating polynomials and introducing Gaussian noise. The polynomial generation process involved random selection of the number of features, ranging between 2 and 5, and the highest exponent for each variable in each term, fluctuating between 1 and 3. Once a polynomial was randomly generated, we produced a predefined number of \(n\) data samples (equivalent to the number of rows in the dataset). These samples were generated by assigning a value between -2 and 2 to each feature, following a uniform distribution. This range was chosen to ensure clear differentiation between symmetric and asymmetric functions, which exhibit different behaviors around zero. Additionally, higher-order polynomials tend to yield values closer to zero in the range of -1 to 1, while larger values between [-2, -1] and [1, 2] allow for rich samples of polynomials, focusing on a relatively small value range to prevent numerical issues such as float overflow. We computed the target feature (\(y\)) for each row based on the polynomial, adding Gaussian noise with a mean value of zero and a standard deviation of 5\% of the original \(y\) value.

\subsection{Real-world Datasets}
For the real-world datasets, we utilized datasets from a recent benchmarking study on automatic ML regression tasks \cite{reg_datasets}. Specifically, we analyzed a total of 20 datasets from nine studies \cite{Huang_et_al, Su_et_al, Atici, Guo_et_al, Koya_et_al, Dunn_et_al, Xiong_et_al, Yin_et_al, Bachir_et_al}. The only modification applied to these datasets was the removal of categorical variables, as this study exclusively focused on numerical variables. This alteration did not impact our ability to evaluate the SRTPOT and SRAK methods, as we compared their results to TPOT and AK models trained on the exact same datasets, specifically using an identical train-test split in each case.

In addition, inspired by \cite{Xie_et_al}, we examined the method's performance in predicting superconducting critical temperatures based on Eliashberg theory. Estimation of superconducting critical temperatures is an extremely active field of research which has drawn extensive scholarly interest \cite{Jose_et_al}. \cite{Xie_et_al} developed an SR model for this purpose using artificially generated \(\alpha^2F\) functions and computed critical temperatures $T_c$ using numerical solutions of the Eliashberg equations. They tested their model on hydride superconductors and found a substantial improvement compared to traditional equations in the field. As \cite{Xie_et_al} had already developed an SR model for this purpose, we extended their findings by training AutoML models either with or without the SR feature based on their work. We decided to use this case due to its physical importance and relatively large dataset.  

\subsection{Evaluation Methodology}
To ensure comprehensive evaluation, we repeated the assessment process 50 times, each time with a different split between the training and validation cohorts, for each dataset, resulting in a total of 1000 runs. For both synthetic and real-world datasets, the training cohort comprised 80\% of the dataset, while the validation cohort contained the remaining 20\%.

To explore the robustness of our proposed method, we conducted two additional experiments. First, we performed a total of 1250 repetitions of the synthetic dataset analysis, varying sample sizes (100, 500, 1000, 5000, 10000) and noise levels (1-5\%). Each configuration consisted of 50 repetitions. Second, we investigated the influence of the number of terms and the non-linearity of the data on our method's performance. Non-linearity was measured in two ways: by dividing the RMSE of a Linear Regression (LR) model by the standard deviation of the target variable and by using $1-R^2$ of the linear regression model trained on the training dataset.

\section{Results}
\label{sec:evolution}

In this section, we present the outcomes of our experiments. We begin by reporting the performance of our proposed method on both synthetic and real-world datasets. Subsequently, we demonstrate the robustness of our approach through various assessments.

\subsection{Performance}
\label{sec:exp1}

\begin{figure}[htp]
    \centering
    \includegraphics[width=14cm]{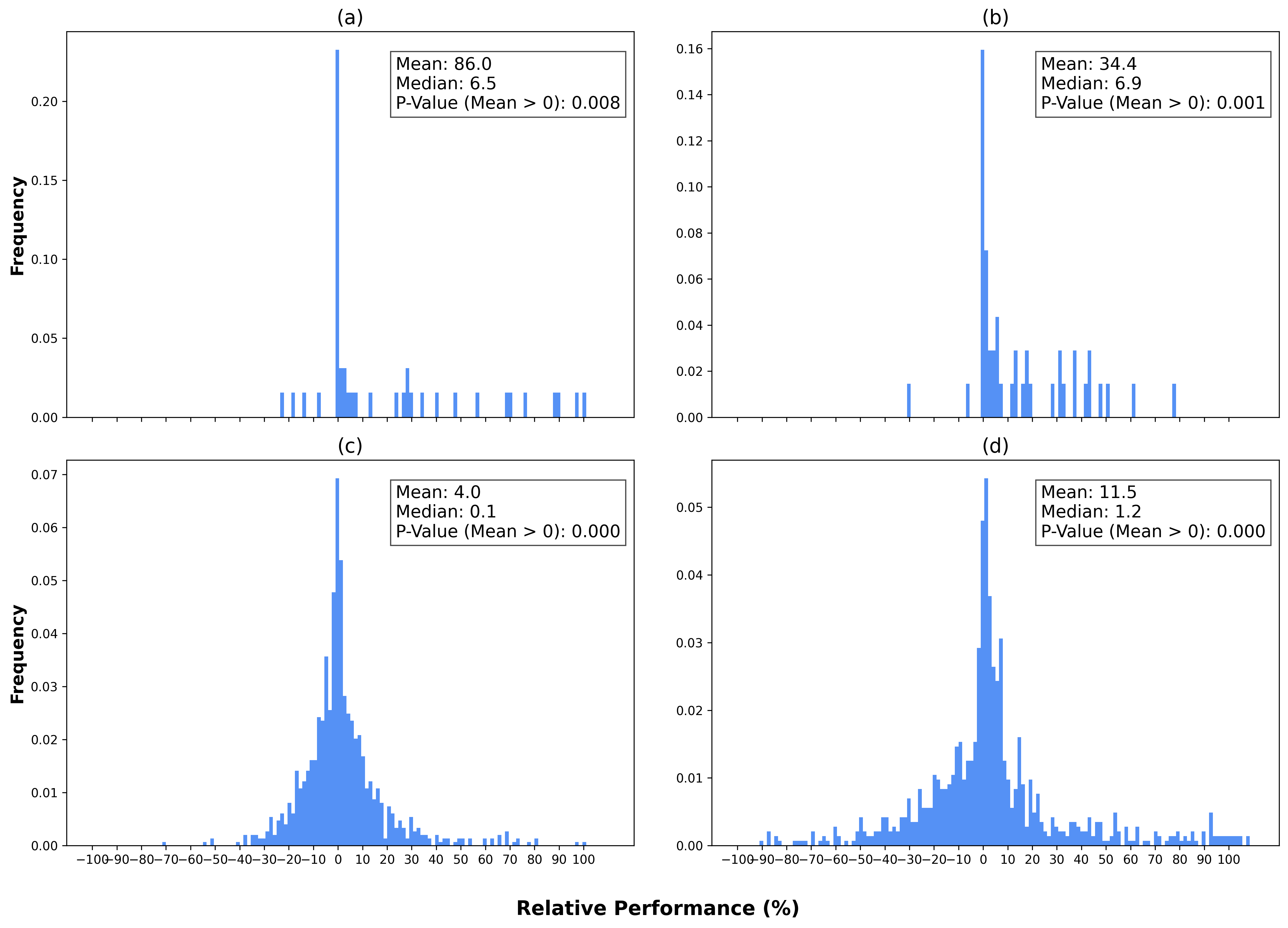}
    \caption{Summary of model performances. (a) and (b) illustrate the relative improvement of the SRTPOT and SRAK models compared to the TPOT and AK models, respectively, in the synthetic datasets. (c) and (d) illustrate the relative improvement of the SRTPOT and SRAK models compared to the TPOT and AK models, respectively, in the real datasets. Values above 100\% are not presented, encompassing approximately 25\% of the observations in subfigures (a) and (b), 1\% of the observations in subfigure (c) and 3\% of the observations in subfigure (d). }
    \label{fig:relative_performance_figure}
\end{figure}

Fig. \ref{fig:relative_performance_figure} provides a comprehensive overview of the model performances. Figs. \ref{fig:relative_performance_figure}a and \ref{fig:relative_performance_figure}b pertain to the synthetic datasets, showcasing the relative improvements achieved by the SRTPOT model over the TPOT model and the SRAK model over the AK model, respectively. These synthetic datasets encompassed 5000 samples in each run and 5\% noise level. Additional datasets are analyzed in Section \ref{sec:exp2}. On average, the TPOT model displayed a mean RMSE of 86.0\% higher than that of the SRTPOT model, with a median difference of 6.5\%. Similarly, the AK model exhibited a mean RMSE of 34.4\% higher than that of the SRAK model, with a median difference of 6.9\%. Notably, the results displayed considerable positive skewness. Fig. \ref{fig:relative_performance_figure} presents values up to 100\%, due to visual convenience, allowing an easier review of the results. The full range plot is provided as supplementary materials. 

Figs. \ref{fig:relative_performance_figure}c and \ref{fig:relative_performance_figure}d detail the results for the real-world datasets. The advantage of the SRTPOT and SRAK models in these datasets is comparatively more modest compared to the synthetic datasets. Nonetheless, it remains substantial in both magnitude and statistical significance (\(p<0.01\)). In particular, the TPOT model was outperformed by 4.0\% (\(p<0.01\)), with a positive median difference slightly above zero. In contrast, the SRAK model outperformed the AK model by 11.5\% (\(p<0.01\)), with a median difference of 1.2\%.

\begin{figure}[htp]
    \centering
    \includegraphics[width=14cm]{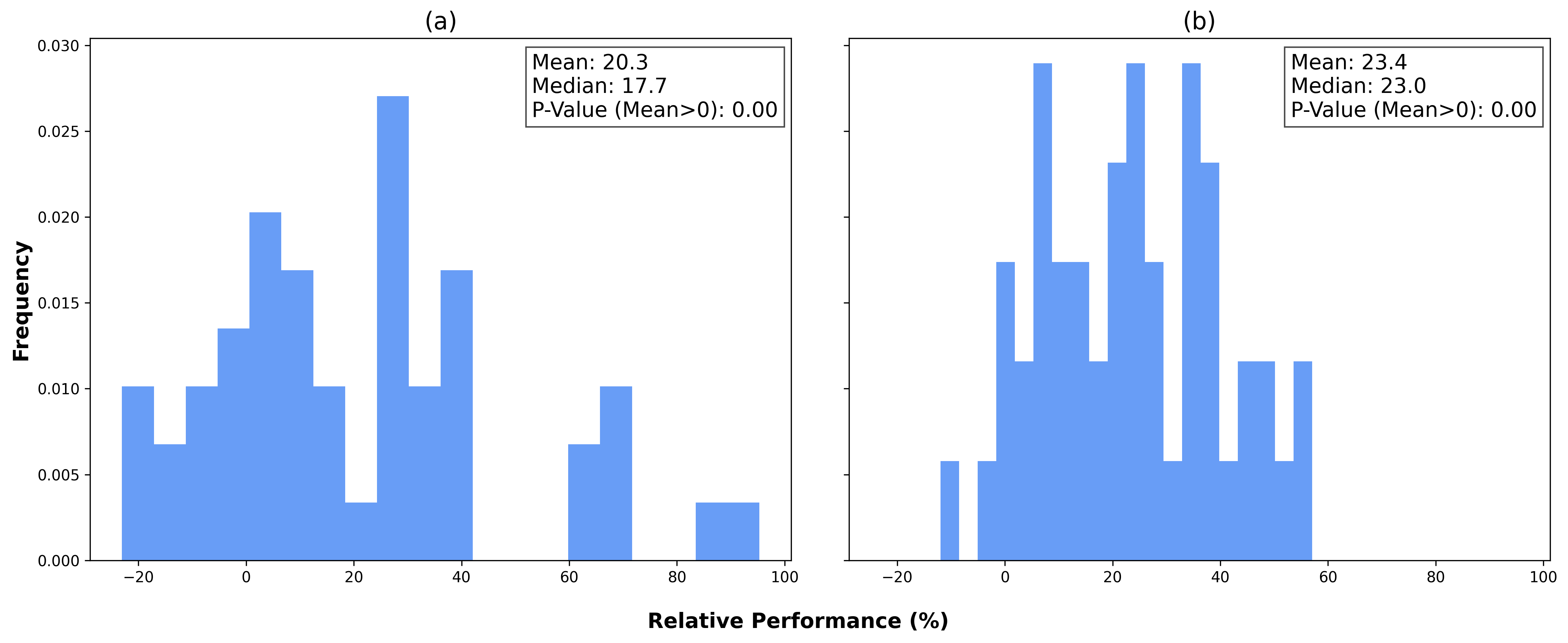}
    \caption{Summary of model performances predicting superconducting critical temperature. (a) and (b) illustrate the relative improvement of the SRTPOT and SRAK models compared to the TPOT and AK models, respectively.}
    \label{fig:Eliashberg}
\end{figure}

Fig. \ref{fig:Eliashberg} presents the relative performances for superconducting critical temperature prediction. The SR-based TPOT and AK models exhibited improvements in mean RMSE by 20.3\% and 23.4\%, respectively. In absolute terms, the SR model developed in \cite{Xie_et_al} obtained an RMSE score of 12.12 degrees. The TPOT and AK models obtained 11.71 and 11.13 degrees RMSE respectively; when integrating the SR model into the TPOT and AK training process, the RMSE scores improved to 9.81 and 9.06 degrees, respectively.

\subsection{Robustness}
\label{sec:exp2}

\begin{figure}[htp]
    \centering
    \includegraphics[width=14cm]{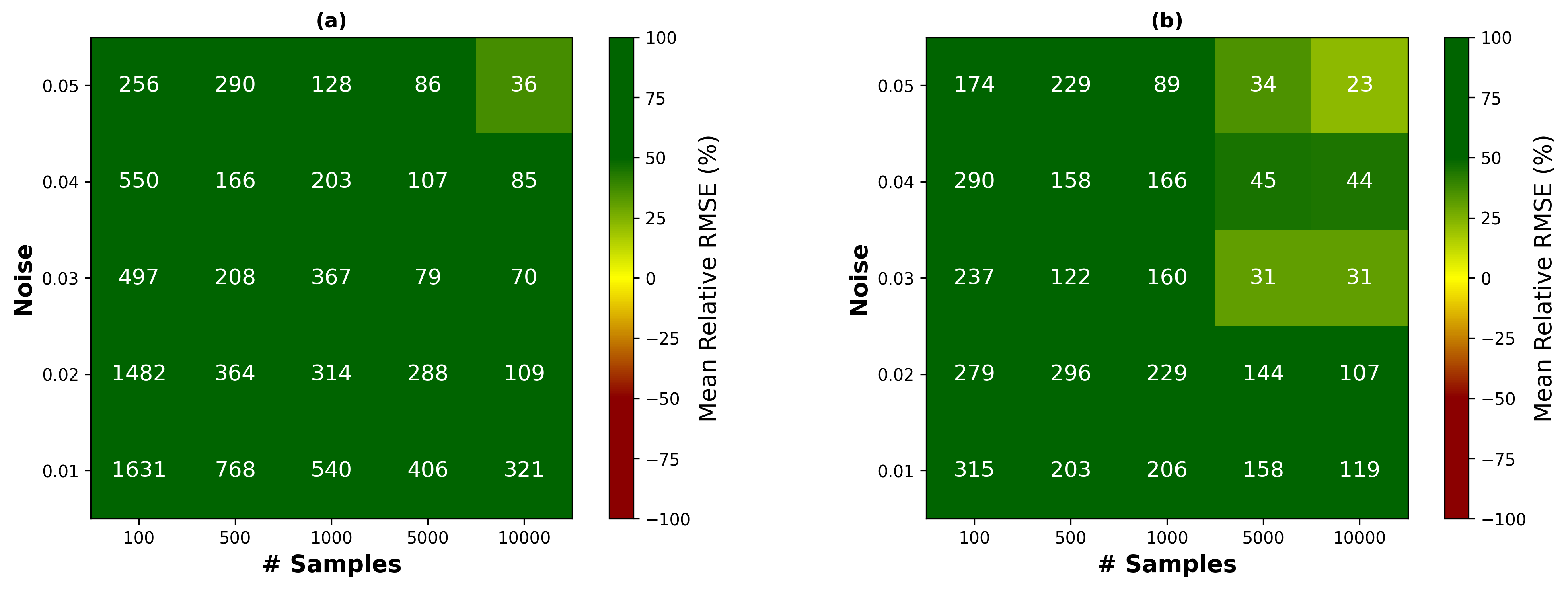}
    \caption{Robustness tests for synthetic data noise and sample size. (a) illustrates the relative improvement of the SRTPOT model compared to the TPOT model. (b) illustrated the relative improvement of the SRAK model compared to the AK model.}
    \label{fig:Robustness tests}
\end{figure}

In addition, Fig. \ref{fig:Robustness tests} presents robustness tests encompassing various sample sizes and noise levels. Across all 25 configurations, which varied in sample size (100, 500, 1000, 5000, 10000) and noise levels (1-5\%) the SRTPOT and SRAK models consistently improved mean RMSE scores. As expected, the most significant improvements were observed in datasets with lower noise levels and smaller sample sizes, aligning with SR's known advantage in such scenarios. Despite a decrease in improvement percentages by 143.7\% (ML) or 25.1\% (DL) for each 1\% increase in noise level, and by 0.05\% (ML) or 0.02\% (DL) for each additional observation, substantial improvements were still observed in datasets with 5\% noise and 10000 observations, showing a mean improvement of 36\% for ML and 23\% for DL, respectively.

\begin{figure}[htp]
    \centering
    \includegraphics[width=14cm]{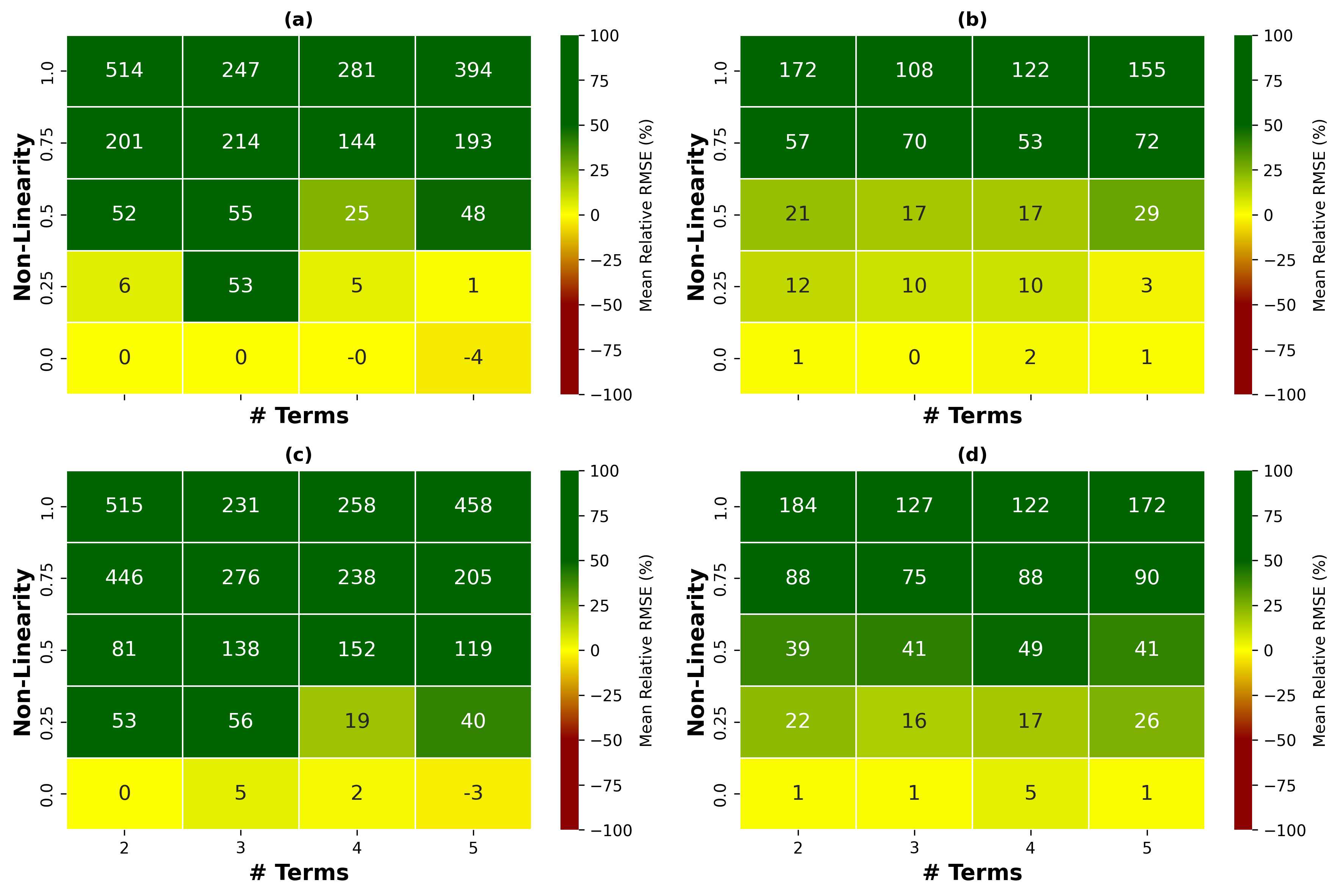}
    \caption{Robustness tests for synthetic data non-linearity. In subfigures (a) and (b), non-linearity is defined as the LR RMSE divided by the standard deviation of the target variable in each dataset. (a) illustrates the relative improvement of the SRTPOT model compared to the TPOT model. (b) illustrates the relative improvement of the SRAK model compared to the AK model. In subfigures (c) and (d), non-linearity is defined as $1-R^2$ of the LR model in each dataset. (c) illustrates the relative improvement of the SRTPOT model compared to the TPOT model. (d) illustrates the relative improvement of the SRAK model compared to the AK model.}
    \label{fig:Robustness tests - non-linearity}
\end{figure}

Moreover, Fig. \ref{fig:Robustness tests - non-linearity} presents robustness tests for the complexity of the generated polynomials. The horizontal axis presents the number of terms in the polynomial expression and the vertical axis presents the non-linearity, measured either with the RMSE score of the LR model divided by the standard deviation of the target variable, or by $1-R^2$ of the LR model. While the number of terms has no statistically significant effect, in all four models it is evident that the contribution of the SR layer is more substantial in non-linear datasets. The Ordinary Least Squares (OLS) coefficients of both robustness tests are summarized in Table \ref{tab:coefficients}.

\begin{table}[ht]
\centering
\caption{Summary of robustness tests}
\begin{threeparttable}
\begin{tabular}{lcccccc}
\textbf{Variable} & \textbf{Model 1} & \textbf{Model 2} & \textbf{Model 3} & \textbf{Model 4} & \textbf{Model 5} & \textbf{Model 6}\\
\hline
Model Type & TPOT & AK 
 & TPOT & AK & TPOT & AK \\
\hline
\# Samples & -0.05 & -0.02  \\
 & (0.01) & (0.00) \\
Noise & -143.7 & -25.1 \\
 & (0.00) & (0.00) \\
\# Terms & & & -10.78 & -0.27 & -17.29 & 0.15 \\
 & & & (0.49) & (0.96) & (0.29) & (0.45) \\
Non-linearity (LR/STD) & & & 356.72 & 132.19 \\
 & & & (0.00) & (0.00) \\
Non-linearity (1-$R^2$) & & & & & 391.50 & 145.05 \\
& & & & & (0.00) & (0.00) \\
\hline
$R^2$ &0.48&0.69&0.76&0.80&0.78&0.88\\
\hline
\end{tabular}
\begin{tablenotes}
\item[Note: P-values are shown in parentheses.]
\end{tablenotes}
\end{threeparttable}
\label{tab:coefficients}
\end{table}

\section{Discussion}
\label{sec:discussion}
In this study, we have introduced a novel approach to FE by incorporating SR into ML and DL regression models. Our findings unequivocally demonstrate the potential of SR-derived features to significantly enhance the predictive performance of data-driven models while reducing the reliance on extensive domain expertise. By enabling models to autonomously evolve mathematical expressions that capture intricate data relationships, SR bridges the gap between interpretability and complexity.

As illustrated in Fig. \ref{fig:relative_performance_figure}, our results show that the proposed SR FE method can statistically improve the average performance of both synthetic and real-world datasets across both ML and DL-based models. As demonstrated in Table \ref{tab:example}, the SR-derived feature improves the TPOT and AK models even though the SR model itself results in an inferior RMSE score. This improvement is independent of the data type or the specific data-driven model used. Moreover, the median value in all cases exceeds 0, indicating that the proposed method enhances the majority of the samples. In more complex real-world datasets, the mean improvement for ML models is 4.0\%, while for DL models, it is 11.5\%. It's important to note that in some cases, many models remain unchanged or experience only slight modifications, as evidenced by the large number of datasets with a relative performance of 0. This outcome is a result of both ML and DL models often performing feature selection, potentially disregarding the features generated by our method if they are found to be non-beneficial.

After establishing the effectiveness of the proposed method, we turned our attention to its robustness. We conducted extensive robustness tests that assessed the contribution of SR in 1250 ML and DL training processes under varying conditions. Based on this data, we computed a two-dimensional sensitivity analysis for the number of samples in a dataset, their noise levels, and the complexity of the regression task (indicated by the number of terms in the mathematical equation) alongside the non-linearity coefficient, as presented in Figs. \ref{fig:Robustness tests} and \ref{fig:Robustness tests - non-linearity}. Based on these results, and as indicated by Table (\ref{tab:coefficients}), we observed a negative correlation between an increase in the number of samples, noise levels, and the relative performance of the proposed method. This observation aligns with the known notion that larger and cleaner datasets are generally easier to handle for data-driven models \cite{data_size_qult}. Nevertheless, even for datasets with 10,000 samples and 5\% Gaussian noise, the proposed method demonstrates a relative average performance increase of 36\% and 23\% for ML and DL models, respectively, reaffirming its robustness to both noise and sample size. Additionally, as datasets become more complex, as depicted in Fig. \ref{fig:Robustness tests - non-linearity}, the proposed method continues to improve (or at least, not worsen) model performance. Notably, for linearly dependent datasets (i.e., with a non-linearity coefficient close to zero), the proposed method does not contribute to model performance. However, even with slight non-linearity, the proposed method demonstrates significant improvement on average.

While our study underscores the potential of SR-derived features, numerous challenges remain. The computational complexity of SR, particularly for larger datasets, necessitates further exploration of efficient optimization strategies \cite{zhou2003extracting}. Furthermore, the applicability of SR may vary depending on dataset characteristics, necessitating investigations into its adaptability to diverse domains and computational properties of the dataset \cite{sr_challanges,sr_benchmark}. Additionally, this study solely employed one SR method - the GPlearn model \cite{gp_model}. Further exploration of other SR models (as proposed in \cite{sr_benchmark}) could yield different results and pave the way for a meta-learning-based approach to select the best SR model based on dataset characteristics \cite{discussion_1}. Moreover, exploring the potential of SR in time series, image analysis, and natural language processing tasks could extend its applicability across various domains.

\section*{Declarations}
\subsection*{Funding}
This research did not receive any specific grant from funding agencies in the public, commercial, or not-for-profit sectors. 

\subsection*{Conflicts of interest/Competing interests}
None. 

\subsection*{Code and Data availability}
The code and data that have been used in this study are publicly available in the research's GitHub repository: \url{https://github.com/AssafS91/Symbolic-Regression-as-Feature-Engineering-Method-for-Machine-and-Deep-Learning-Regression-Tasks}.

\subsection*{Acknowledgement}
We express our gratitude to Amir Dalal for his invaluable support in managing the data related to superconducting critical temperatures. T.L. wishes to thank Alex Liberzon for encouraging him to pursue this line of work.

\subsection*{Author Contribution}
Assaf Shmuel: Conceptualization, Methodology, Software, Formal Analysis, Investigation, Visualization, Writing - Original Draft. \\  
Oren Gilckman: Conceptualization, Investigation, Supervision, Writing - Review \& Editing. \\
Teddy Lazebnik: Conceptualization, Methodology, Investigation, Validation, Project Administration, Supervision, Writing - Original Draft, Writing - Review \& Editing. \\ 
 
\bibliography{biblio}
\bibliographystyle{unsrt}

\end{document}



\title{\Large Symbolic Regression as Feature Engineering Method for Machine and Deep Learning Regression Tasks}

\author{Assaf Shmuel$^{1*}$, Oren Glickman$^{1}$, Teddy Lazebnik$^{2}$\\
\(^1\) Department of Computer Science, Bar Ilan University, Ramat Gan, Israel\\
\(^2\) Department of Cancer Biology, Cancer Institute, University College London, London, UK\\
\(*\) Corresponding author: assafshmuel91@gmail.com

}

\date{ }

\maketitle 

\maketitle \thispagestyle{empty}

\pagestyle{myheadings} \markboth{Draft:  \today}{Draft:  \today}
\setcounter{page}{1}

\section*{Appendix}

\subsection*{Pseudo-code of the process}
Below, we provide a pseudo-code of the proposed method. 

\begin{algorithm}[t]
\label{algo:gen}
	\caption{Pseudocode of SRTPOT and SRAK algorithms} \label{algo}
	\begin{algorithmic}[1]
	    \STATE $\text{\textbf{Input: }} \text{dataset ($x_1...x_n$,y) }  , \text{scoring metric} , \text{SR, TPOT, and AK hyperparameters}$
	    \STATE $\text{\textbf{Output: }} \text{relative TPOT/SRTPOT performance, relative AK/SRAK performance } $
	    \STATE $X_{train},X_{test},y_{train},y_{test} \Leftarrow \text{random balanced train-test split }$
     
         \STATE $\text{\textbf{function } \text{train\_SR}}(X_{train}, X_{test}, y_{train}, \text{SR hyperparameters})$
            \STATE $SR\_model \Leftarrow \text{Train SR model using} X_{train} \text{ and } y_{train}$
        \STATE $x_{SR} \Leftarrow \text{SR\_model.predict(} X_{test} \text{)} $
        \STATE $\text{return } X_{SR}_{train} \text{, } X_{SR}_{test}$
       \STATE \textbf{end function}
         \STATE $\text{\textbf{function } \text{train\_TPOT}}(X_{train}, X_{test},y_{train}, y_{test} \text{, TPOT hyperparameters})$
        \STATE $TPOT\_model \Leftarrow \text{Train TPOT model using} X_{train} \text{ and } y_{train}$
        \STATE $y_{predicted} \Leftarrow \text{TPOT\_model.predict(} X_{test} \text{)} $
        \STATE $score \Leftarrow \text{Calculate error between } y_{predicted} \text{ and } y_{test}$
        \STATE return score
         \STATE \textbf{end function}
         \STATE $\text{\textbf{function } \text{train\_AK}}(X_{train}, X_{test},y_{train}, y_{test} \text{, AK hyperparameters})$
        \STATE $TPOT\_model \Leftarrow \text{Train AK model using} X_{train} \text{ and } y_{train}$
        \STATE $y_{predicted} \Leftarrow \text{AK\_model.predict(} X_{test} \text{)} $
        \STATE $score \Leftarrow \text{Calculate error between } y_{predicted} \text{ and } y_{test}$
        \STATE return score
         \STATE \textbf{end function}

        \STATE $TPOT\_score \Leftarrow train\_TPOT \text{(} X_{train}\text{, } X_{test} \text{, } y_{train}\text{, } y_{test} \text{, TPOT hyperparameters)}$
        \STATE $AK\_score \Leftarrow train\_AK \text{(} X_{train}\text{, } X_{test} \text{, } y_{train}\text{, } y_{test} \text{, AK hyperparameters)}$
        \STATE $x_{SR}_{train} \text{, } x_{SR}_{test} \Leftarrow train\_SR \text{(}X_{train}\text{, }y_{train} \text{, SR hyperparameters)}$
        \STATE $X_{train\_SR} \Leftarrow X_{train} \text{, } x_{SR}_{train}$
        \STATE $X_{test\_SR} \Leftarrow X_{test} \text{, }x_{SR}_{test}$
        \STATE $SRTPOT\_score \Leftarrow \text{train\_TPOT}(X_{train\_SR}, X_{test\_SR},y_{train}, y_{test} \text{, TPOT hyperparameters})$
        \STATE $SRAK\_score \Leftarrow \text{train\_AK}(X_{train\_SR}, X_{test\_SR},y_{train}, y_{test} \text{, AK hyperparameters})$
        \STATE $TPOT/SRTPOT\_score \Leftarrow TPOT\_score \text{/} SRTPOT\_score$
        \STATE $AK/SRAK\_score \Leftarrow AK\_score \text{/} SRAK\_score$
        \STATE $\text{return } TPOT\_score/SRTPOT\_score\text{, } AK\_score/SRAK\_score$
\end{algorithmic}
\end{algorithm}

\subsection*{Feature selection}
The primary objective of the F approach is to identify a subset from the initial feature set, optimizing a designated utility function. FS algorithms can be roughly divided into three main methodologies: Filters, Embedding, and Wrappers \cite{intro_2}. Filter FS algorithms function as an initial processing step, whereby features are ranked based on their attributes, and the most highly ranked features are subsequently chosen to reduce the dimensionality of the dataset \cite{intro_2}. These ranking criteria encompass various statistical measures that capture the intrinsic properties of the data, including measures of distance, dependency, consistency, and correlation between individual features or feature groups and the target variable, which is typically the class being learned. Conversely, Embedding FS algorithms are executed in conjunction with a particular machine learning (ML) algorithm, strategically conducting feature selection during the model training process \cite{intro_3}. Meanwhile, Wrapper FS algorithms operate on the premise of maximizing the ML model's performance by encompassing a search algorithm that iteratively seeks the optimal feature subset, thereby ensuring the highest model performance.

Numerous Filter FS algorithms have been developed, each with distinct methodologies. For instance, Remove Low Variance \cite{intro_2} ascertains feature ranking based on variance, selectively eliminating features with variances below a predetermined threshold. Chi-square \cite{chi_square} hinges on the Chi-square test to gauge the association between independent features and the dependent (target) feature, favoring features with higher dependence on the target variable. Information gain \cite{information_gain} employs an entropy-based approach, ranking features by their contribution to the output data while discarding low-impact features below a predetermined threshold.

For the embedding FS algorithms, two primary sub-categories can be identified: tree-based and coefficients-based models. The tree-based FS approach computes the average contribution of each feature towards class classification. An example of this is the scikit-learn Python library (\textit{sklearn}), which employs the mean decrease impurity metric (i.e., the GINI index) for tree-based ML models like decision trees and random forests to gauge feature importance \cite{gini,dt,rf}. In contrast, coefficients-based models, such as Lasso and Support Vector Machine (SVM) algorithms, yield coefficient vectors for a particular family of functions (typically linear or polynomial). If the sum of coefficients for a given feature is zero, it implies that the feature had no discernible impact on model optimization during the training phase and can be pruned.

\bibliography{biblio}
\bibliographystyle{unsrt}